\documentclass[preprint,authoryear,11pt]{elsarticle}

\usepackage[T1]{fontenc}
\usepackage[utf8]{inputenc}
\usepackage{mathptmx}    
\usepackage{microtype}
\usepackage{amsmath}
\usepackage{amssymb}
\usepackage{graphicx}
\usepackage{booktabs}
\usepackage{url}
\usepackage[hidelinks]{hyperref}

\makeatletter
\def\ps@pprintTitle{%
  \let\@oddhead\@empty
  \let\@evenhead\@empty
  \def\@oddfoot{\centerline{\thepage}}%
  \let\@evenfoot\@oddfoot}
\makeatother

\journal{arXiv}

\begin{document}

\begin{frontmatter}

\title{Analyzing Public Discourse on Urbanism: Topic Clustering, Sentiment
Analysis and Retrieval-Augmented Generation using YouTube Comments}


\author[inst1]{Jakob Morales\fnref{equal}}

\author[inst1]{Monica Hegde\fnref{equal}}

\author[inst1]{and Fayeq Jeelani Syed\corref{cor1}}

\ead{faysyed@iu.edu}

\fntext[equal]{Monica Hegde and Jakob Morales contributed equally to this work.}

\cortext[cor1]{Corresponding author.}

\affiliation[inst1]{
    organization={Luddy School of Informatics, Computing and Engineering, Indiana University},
    city={Indianapolis},
    state={Indiana},
    country={USA}
}

\begin{abstract}
Online discourse about urban issues---walkability, cycling infrastructure,
public transit, housing density, and street safety---is voluminous but
unstructured, and existing city-evaluation tools capture none of it. We
present a pipeline and conversational system that combines geographic
entity resolution, topic modeling, sentiment analysis, and
Retrieval-Augmented Generation (RAG) over 22{,}788 chunks of YouTube
transcripts and comments spanning 309 North American cities. Beyond the
system itself, our contribution is a set of measurements about what
happens when standard NLP components meet short, informal, geographically
ambiguous text. A Twitter-tuned RoBERTa classifier outperforms a VADER
lexicon baseline by 12.6 macro-F1 points (0.589 vs.\ 0.464; McNemar
$p = 0.0001$), but both models collapse on the neutral class, which
dominates urbanist comment traffic; annotators disagree on the same class
($\kappa = 0.53$). Dense retrieval beats a TF-IDF baseline at every cutoff
(P@5 0.790 vs.\ 0.560), and video-level relevance proxies understate
chunk-level precision by a wide margin (0.660 vs.\ 0.94 under human
rating). For groundedness evaluation we find BERTScore unusable when a
multi-sentence generated summary is compared against a single short
comment---scores are nearly flat regardless of relevance---and show that
ROUGE-1-based groundedness is a paraphrase-driven lower bound rather than
a hallucination rate. These findings generalise beyond the urbanist domain
to any RAG system built over short user-generated documents.
\end{abstract}


\end{frontmatter}

\section{Introduction}

Urban discourse covering walkability, cycling infrastructure, public
transit, housing density, and street safety has entered mainstream public
conversation. Google Trends data shows increased search interest in terms
such as \textit{walkability}, \textit{car dependency}, \textit{housing
density}, \textit{safe streets}, and \textit{public transit}
\citep{gtrends_urbanism}. YouTube has emerged as a
significant platform for this discourse, where urbanist channels generate
large volumes of transcripts and comments in which users discuss and
evaluate specific cities. Existing tools for evaluating cities, such as
Walk Score, rely on quantitative proximity-to-amenity indicators
\citep{walkscore_methodology} and capture neither subjective public
sentiment \citep{manaugh2011validating} nor the qualitative reasoning
expressed in online discussion.

\paragraph{Task} Given a natural language query (e.g.\ \textit{``Is
Chicago walkable?''}), the system must retrieve relevant text segments,
associate them with the correct city, and produce a response grounded in
the retrieved evidence, returning a generated answer together with the
detected city, identified topic aspects, a sentiment summary, and
supporting passages. This requires resolving implicit and informal
geographic references, classifying text into urbanist categories, and
estimating sentiment per segment. Unlike standard document retrieval, each
retrievable unit is a fragmented, sentiment-bearing comment rather than a
coherent document---the regime in which retrieval precision is known to
degrade \citep{gao2023ragsuvey}.

\paragraph{Contributions}
\begin{enumerate}
  \item An urbanist YouTube corpus of 218{,}177 raw transcript/comment
  records across 743 videos, reduced by length filtering (169{,}548),
  supervised topic-relevance filtering (93{,}651 on-topic, 55.2\%), and
  chunking to 22{,}788 retrieval chunks over 309 geographically resolved
  cities, with geographic, topic, and sentiment annotations.
  \item Per-city, per-topic sentiment profiles across six urbanist
  dimensions, and a RAG conversational system that grounds natural
  language answers in them.
  \item A \textit{topic-versus-video} sentiment coding rule for
  platform-native user-generated text, which separates opinion about the
  subject matter from opinion about the creator, together with an error
  analysis showing that the neutral class is the dominant failure mode for
  both lexicon and transformer classifiers in this register.
  \item Three transferable evaluation findings for RAG over short
  user-generated text: (i) BERTScore is uninformative under the length
  mismatch between a generated summary and a single short comment;
  (ii) ROUGE-1 groundedness is a lower bound driven by paraphrase, not a
  hallucination rate; and (iii) video-level relevance proxies
  substantially understate chunk-level retrieval precision.
\end{enumerate}

\section{Related Work}

\paragraph{Sentiment on social media} The VADER lexicon
\citep{hutto2014vader} was designed for social media text and remains a
widely used baseline. \citet{nahas2024classifiers} show that transformer
models outperform lexicon methods on informal YouTube comment sentiment.
Neither line of work addresses domain-specific urbanist language---terms
such as \textit{bikeability}, \textit{upzoning}, or \textit{stroad} carry
polarity that general-purpose models are not calibrated for---nor
associates sentiment with geographic location. We build on the transformer
approach and quantify where it fails in this register.

\paragraph{Topic modeling} BERTopic \citep{grootendorst2022bertopic}
combines transformer embeddings with HDBSCAN clustering
\citep{mcinnes2017hdbscan} and class-based TF-IDF, producing more coherent
topics than LDA on short, noisy text. The most structurally similar prior
work is \citet{zapletal2023social}, who apply BERTopic and sentiment
analysis to municipal tweets and aggregate per-topic civic sentiment at
the city level. Their system covers a single European city, uses TextBlob
for sentiment, and provides no mechanism for querying the resulting
profiles.

\paragraph{Geoparsing} Associating text with places is a long-studied
problem in its own right: toponym recognition and resolution
\citep{leidner2008toponym}, the pervasive ambiguity of place names
\citep{gritta2018melbourne}, event-oriented geoparsing pipelines
\citep{halterman2017mordecai}, and evaluation practice for geoparsers
\citep{gritta2020pragmatic}. Our NER-plus-fuzzy-matching approach is
deliberately lightweight rather than novel; we report its failure modes
rather than claiming an advance over this literature.

\paragraph{Retrieval-augmented generation} \citet{lewis2020rag}
established the RAG paradigm, and \citet{gao2023ragsuvey} document that
retrieval precision degrades on short, semantically sparse documents. RAG
has been applied to education, legal question answering, and customer
service; we are not aware of prior work applying it over geographically
indexed social media for place-based discourse querying. The integration
of sentiment aggregation, topic structure, and retrieval over
geographically grounded urban discourse is the contribution of this work.

\section{Method}

\begin{figure*}[t]
\centering
\includegraphics[width=\textwidth]{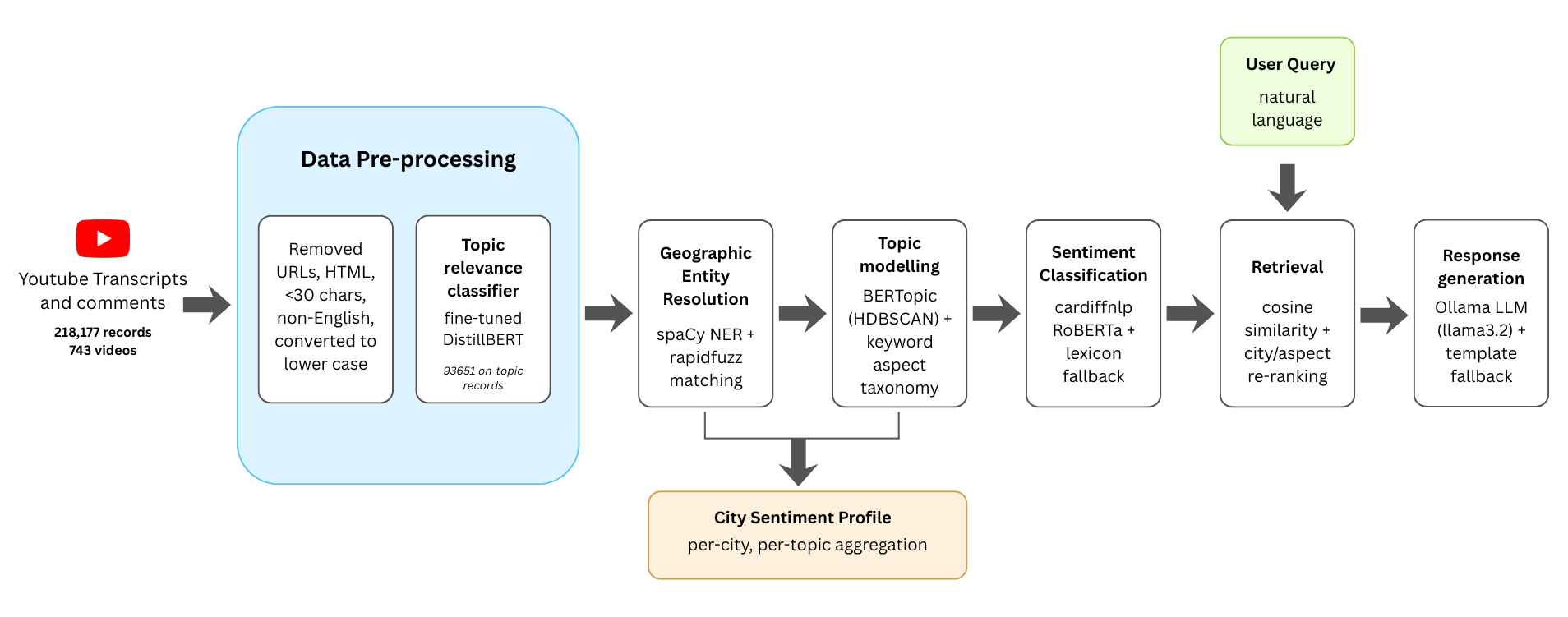}
\caption{System architecture: preprocessing (data cleaning and
topic-relevance classification) followed by geographic entity resolution,
topic modeling, sentiment classification, document indexing, retrieval,
and response generation, with the specific method used at each stage.}
\label{fig:architecture}
\end{figure*}

The system is a multi-stage pipeline: preprocessing and topic-relevance
filtering, geographic entity resolution, topic modeling, sentiment
classification, indexing, retrieval, and generation
(Figure~\ref{fig:architecture}). Library versions and API details are in
\ref{app:impl}.

\subsection{Data Collection and Preprocessing}
Videos were selected by keyword search over walkability, public transit,
cycling infrastructure, housing density, street design, and traffic and
pedestrian safety, across North American cities. Transcripts were obtained
with the YouTube Transcript API and comments via the YouTube Data API
(\texttt{commentThreads}). Preprocessing removes URLs, HTML, excessive
whitespace, and repeated characters; records under 30 characters are
dropped; text is lowercased; non-English comments are discarded; and stop
words---extended with country names, to prevent country-level geographic
bias---are removed.

\subsection{Topic-Relevance Filtering}
\label{sec:relevance}
A large fraction of comments on urbanist videos concern the video rather
than any urban topic. We therefore fine-tune a DistilBERT classifier
\citep{sanh2019distilbert} to a binary on-topic/off-topic decision, where
\textit{off-topic} covers video praise, creator commentary, and discussion
unrelated to any urban subject. Applying it to the 169{,}548
length-filtered records retains 93{,}651 on-topic records (55.2\%). Named
entity resolution and chunking over the on-topic corpus yield 22{,}788
retrieval chunks spanning 309 geographically resolved cities. We do not
report held-out performance for this classifier; the consequences of this
omission are discussed in Section~\ref{sec:limitations}.

\subsection{Geographic Entity Resolution}
spaCy's NER model \citep{honnibal2020spacy} identifies \texttt{GPE}
entities, which are matched against a US/Canada city gazetteer using
rapidfuzz's weighted-ratio scorer with an acceptance threshold of 95/100.
An alias dictionary is checked first as a fast path, mapping informal
references such as \textit{nyc} to New York City. For corpus chunks, the
primary city is the highest-weight entry in a precomputed mention
distribution. This is more robust to the informal references common in
YouTube comments than exact matching, but inherits the ambiguity problems
documented in the geoparsing literature.

\subsection{Topic Modeling}
BERTopic is fit over all 22{,}788 chunks, reusing the 384-dimensional
all-MiniLM-\allowbreak L6-v2 embeddings \citep{reimers2019sbert} computed for the
retrieval index. HDBSCAN uses a minimum cluster size of 10; topic labels
are the three highest-weighted terms per topic; unclustered documents are
assigned to an outlier class. In parallel, a keyword taxonomy of six
aspect categories---transit, walkability, bikeability, car dependency,
safety, and a density/land-use signal covering housing density, sprawl,
and zoning---is applied as a supplementary labeling layer. Multi-word
phrases are matched by substring, single words by word-boundary regex, and
a chunk may receive any subset of the six labels. Both label types are
stored per chunk and used at retrieval and generation time.

\subsection{Sentiment Classification}
Polarity is classified with
\texttt{cardiffnlp/\allowbreak twitter-roberta-\allowbreak base-sentiment-\allowbreak latest}
\citep{liu2019roberta, loureiro2022timelms}, a RoBERTa
model fine-tuned on approximately 124 million tweets, truncating to 512
tokens. The corpus-level distribution is 44.8\% negative, 31.6\% neutral,
23.6\% positive. A lexicon fallback using 27 positive and 27 negative seed
words is retained for environments where the transformer cannot be loaded;
it is not used in any reported result.

\subsection{Indexing, Retrieval, and Generation}
All chunks are indexed with FAISS \texttt{IndexFlatIP}
\citep{johnson2019faiss} over L2-normalised vectors, giving exact cosine
search on CPU. A TF-IDF sparse index (20{,}000 features, unigrams and
bigrams, sublinear TF, min-DF 2) serves both as a fallback and as the
retrieval baseline in Section~\ref{sec:retrieval-eval}.

At query time the system parses the query for a target city and aspect set
using the same resolution and keyword procedures, encodes it densely, and
takes the top 200 candidates by cosine similarity. If a city is detected,
the pool is filtered to that city provided at least three results remain.
If aspects are detected, an aspect-overlap score weighted at 0.2 is added
to the similarity score for re-ranking. The top 12 chunks are passed to
generation, formatted with source type, resolved city, aspects, topic
label, sentiment, and up to 400 characters of text, prepended by the city
sentiment profile where available. The prompt instructs the model to
answer solely from the provided evidence, summarise sentiment trends,
surface recurring themes, and state when evidence is insufficient; output
is capped at 600 tokens. Generation uses a locally hosted Llama 3.2
\citep{grattafiori2024llama3} served by Ollama. Where the model server is
unavailable, a template-based fallback aggregates sentiment counts and
constructs a response from representative evidence passages; because this
fallback copies evidence spans directly, it inflates surface-overlap
metrics, and its effect on the reported groundedness figures is discussed
in Section~\ref{sec:limitations}.

\subsection{Training Regime}
Only the topic-relevance classifier of Section~\ref{sec:relevance} is
trained by us. BERTopic is fit in a single unsupervised pass over the full
corpus with no train/test split, as topic modeling is exploratory and has
no ground truth. The sentence encoder and sentiment classifier are used
with pre-trained weights, and the language model is prompted zero-shot.

\section{Evaluation Setup}

The pipeline produces two outputs, evaluated separately. \textbf{City
profiles} are assessed via sentiment classification performance
(precision, recall, macro-F1, MCC against a manually labeled validation
set, with VADER as baseline and McNemar's test plus bootstrap confidence
intervals for significance) and topic coherence ($C_v$). \textbf{Chatbot
responses} are assessed via retrieval quality (Precision@$k$/Recall@$k$,
$k \in \{5,10\}$, dense vs.\ TF-IDF, with video-level judgments validated
by a chunk-level human rating) and groundedness (ROUGE-1 recall
\citep{lin2004rouge} and sentence-embedding cosine similarity between each
response and its 12 retrieved chunks) over a held-out set of 20 urbanist
queries spanning 9 cities and all 6 aspect categories.

\section{Results}

\subsection{Corpus Statistics}
The raw corpus contains 218{,}177 records across 743 unique video IDs
(217{,}507 comments, 670 transcripts). Removing records below the
30-character threshold discards 48{,}629, leaving 169{,}548; relevance
filtering retains 93{,}651 (55.2\%); chunking and geographic resolution
yield 22{,}788 chunks over 309 cities. The corpus-level sentiment
distribution is 44.8\% negative, 31.6\% neutral, 23.6\% positive,
reflecting the substantial share of traffic- and pedestrian-safety
discourse, where comments more often describe hazards, near-misses, and
criticism of road design than the more general urbanist discussion of
walkability or transit quality.

\subsection{City Sentiment Profiles}
\label{sec:city-profiles}
Per-city profiles sum each chunk's topic weights and sentiment label and
normalise both to proportions. Table~\ref{tab:cityprofiles} shows the
eight cities with the largest chunk coverage.

\begin{table}[htbp]
\centering
\footnotesize
\setlength{\tabcolsep}{4pt}
\caption{City sentiment profiles for the eight highest-coverage cities:
three highest-weighted topic dimensions (normalized proportion of that
city's aggregated topic weight) and normalized sentiment share.
$^{\dagger}$New York and Manhattan are resolved as distinct entities and
are reported separately; $^{\ddagger}$``London'' is matched against the
US/Canada gazetteer and is not disambiguated from London, UK.}
\label{tab:cityprofiles}
\begin{tabular}{@{}llrrr@{}}
\toprule
\textbf{City} & \textbf{Top dimensions (weight)} & \textbf{Neg.} & \textbf{Neu.} & \textbf{Pos.} \\
\midrule
Houston   & car\_dependency (.361), walkability (.303), transit (.170) & 57.2\% & 26.4\% & 16.4\% \\
New York$^{\dagger}$  & car\_dependency (.535), transit (.191), walkability (.111) & 51.7\% & 33.9\% & 14.4\% \\
Chicago   & transit (.347), car\_dependency (.332), walkability (.154) & 40.4\% & 25.4\% & 34.3\% \\
Manhattan$^{\dagger}$ & car\_dependency (.546), transit (.206), walkability (.110) & 38.3\% & 46.9\% & 14.8\% \\
Seattle   & transit (.484), car\_dependency (.264), walkability (.092) & 41.4\% & 27.6\% & 30.9\% \\
London$^{\ddagger}$    & car\_dependency (.392), transit (.283), walkability (.129) & 40.9\% & 36.7\% & 22.4\% \\
Montreal  & transit (.289), bikeability (.268), car\_dependency (.266) & 25.6\% & 38.4\% & 36.0\% \\
Portland  & transit (.322), car\_dependency (.274), bikeability (.210) & 43.1\% & 25.8\% & 31.1\% \\
\bottomrule
\end{tabular}
\end{table}

The profiles are city-differentiated rather than uniform. Car dependency
dominates the topic weight for Houston and the New York entities,
consistent with car-oriented street-safety and traffic discourse, whereas
Montreal and Portland split more evenly between transit and bikeability.
Sentiment varies substantially: Montreal has the lowest negative share
(25.6\%) and highest positive share (36.0\%), while Houston and New York
are most negative (57.2\% and 51.7\%). Manhattan has the highest neutral
share (46.9\%), plausibly reflecting the volume of factual,
incident-reporting comments---describing a specific crash or
intersection---in the safety-focused portion of the corpus.

Two caveats apply. First, \textit{New York} and \textit{Manhattan} are
resolved as distinct entities: the resolver operates on surface mentions
and has no containment hierarchy, so Manhattan-specific discourse is not
folded into New York. Second, \textit{London} is matched against a
US/Canada gazetteer, so mentions of London, UK are absorbed into a North
American entry rather than rejected. Both are entity-resolution
limitations of the lightweight geoparsing approach rather than properties
of the discourse.

\subsection{Sentiment Classification}
The validation set is 300 comments drawn as a stratified random sample,
balanced across short (30--100 chars), medium (100--300), and long (300+)
buckets, each labeled positive, negative, or neutral. A coding rule was
applied throughout: \textbf{sentiment is coded with respect to the urban
topic under discussion, not the video or creator}. Comments such as
``great video!'' are neutral, because they express no opinion about any
city, transit system, or urban environment. The distribution is 122
neutral (40.7\%), 103 negative (34.3\%), 75 positive (25.0\%). This
differs from the corpus-level distribution (44.8\% negative) because the
validation set is stratified by length rather than sampled
proportionally, over-representing short comments.

Two annotators independently labeled all 300 comments under this rule.
Agreement is Cohen's $\kappa = 0.53$ (69.7\% raw), i.e.\ moderate
\citep{landis1977kappa}. Disagreement concentrates on the positive class:
of 75 comments the first annotator called positive, the second called 22
neutral and 12 negative---the same positive-class ambiguity the automated
classifiers exhibit, and consistent with urbanist praise being expressed
in domain-specific phrasing that reads as neutral or backhanded. Model
results are scored against a single gold label set; the second annotation
is used solely to estimate agreement.

\begin{table}[htbp]
\centering
\small
\caption{Sentiment classification on 300 manually labeled urbanist
YouTube comments. MCC = Matthews Correlation Coefficient.}
\label{tab:sentiment}
\begin{tabular}{@{}lcccc@{}}
\toprule
\textbf{Model} & \textbf{P} & \textbf{R} & \textbf{F1 (macro)} & \textbf{MCC} \\
\midrule
VADER (lexicon)          & 0.510 & 0.510 & 0.464 & 0.261 \\
Transformer (cardiffnlp) & 0.588 & 0.601 & 0.589 & 0.393 \\
\bottomrule
\end{tabular}
\end{table}

\begin{table}[htbp]
\centering
\small
\caption{Per-class F1 and recall on the validation set.}
\label{tab:perclass}
\begin{tabular}{@{}lcccccc@{}}
\toprule
& \multicolumn{3}{c}{\textbf{F1}} & \multicolumn{3}{c}{\textbf{Recall}} \\
\cmidrule(lr){2-4}\cmidrule(l){5-7}
\textbf{Model} & Pos. & Neg. & Neu. & Pos. & Neg. & Neu. \\
\midrule
VADER       & 0.50 & 0.55 & 0.35 & 76.0\% & 52.4\% & 24.6\% \\
Transformer & 0.54 & 0.71 & 0.51 & 54.7\% & 79.6\% & 45.9\% \\
\bottomrule
\end{tabular}
\end{table}

VADER reaches 0.47 accuracy, the transformer 0.60. The transformer gains
12.6 macro-F1 points and 13.2 MCC points (Table~\ref{tab:sentiment}), with
per-class detail in Table~\ref{tab:perclass}.

\subsection{Significance of the Sentiment Comparison}
\label{sec:sentiment-significance}
Both models are scored against the same 300 gold labels, giving paired
per-comment predictions. McNemar's test on the $2\times2$ table of
(VADER correct, transformer correct) pairs gives $\chi^2 = 14.88$,
$p = 0.0001$: the disagreement pattern is asymmetric, so the transformer
corrects VADER's errors significantly more often than the reverse. A
bootstrap 95\% confidence interval on the macro-F1 difference
(10{,}000 resamples) is $[0.067, 0.185]$, excluding zero.

\subsection{Retrieval Evaluation}
\label{sec:retrieval-eval}
Retrieval is evaluated over the 20 held-out queries, comparing dense FAISS
retrieval against TF-IDF cosine similarity. Relevance judgments are
video-level: for each query, any video containing at least one chunk
matching the expected city and topic category counts as relevant, then
spot-checked by hand. All 20 queries had at least one relevant video.

\begin{table}[htbp]
\centering
\small
\caption{Retrieval Precision@$k$ and Recall@$k$, dense (FAISS) vs.\
TF-IDF baseline, averaged across 20 held-out queries. Judgments are
video-level.}
\label{tab:retrieval}
\begin{tabular}{@{}lcccc@{}}
\toprule
\textbf{Method} & \textbf{P@5} & \textbf{R@5} & \textbf{P@10} & \textbf{R@10} \\
\midrule
Dense (FAISS)   & 0.790 & 0.230 & 0.660 & 0.372 \\
TF-IDF baseline & 0.560 & 0.183 & 0.545 & 0.302 \\
\bottomrule
\end{tabular}
\end{table}

Dense retrieval wins at both cutoffs on both metrics, with the largest gap
at $k=5$ (0.790 vs.\ 0.560), where a RAG context window is most sensitive
to ranking quality.

\paragraph{Chunk-level human rating} To assess how conservative the
video-level proxy is, the top 10 chunks retrieved by the dense system for
each of the 20 queries (200 chunks) were rated individually for relevance
by one author, who was not blind to the system under evaluation. Mean
Precision@10 is 0.94 (188/200), against 0.660 under the video-level proxy.
Nineteen queries scored $\geq 0.9$; the outlier is \textit{``Is Edmonton
bikeable?''} at 0.5, where retrieved chunks came from multi-city cycling
threads about Winnipeg, Ottawa, and Calgary. The video-level proxy is
therefore conservative for this system, and its main failure is precisely
the multi-city-video problem this exercise surfaces. Because the TF-IDF
baseline was not rated under the same protocol, this figure characterises
the dense system only and is not a comparative result.

\subsection{Topic Coherence}
BERTopic identifies 195 non-outlier topics. $C_v$ coherence over the top
10 words per topic, using gensim's sliding-window estimator, is 0.517,
indicating that each topic's top terms co-occur substantially more often
than chance would predict. We do not fit a comparable LDA model on this
corpus, so this figure is reported without a within-corpus reference
point.

\subsection{Chatbot Groundedness}
\label{sec:groundedness}
For each query, ROUGE-1 recall and sentence-embedding cosine similarity
are computed between the generated response and each of the 12 retrieved
chunks. BERTScore \citep{zhang2020bertscore} was evaluated first as the
semantic complement to ROUGE-1 and found unusable here: raw BERTScore
(RoBERTa-large token matching) was nearly flat ($\sim$0.83--0.86)
regardless of actual relevance when a long, multi-sentence generated
summary is compared against a single short informal comment, and its
rescaled variant collapsed well-grounded pairs to near-zero or negative
scores, because its baseline is not calibrated for this length-mismatched,
informal-text comparison. Sentence-embedding cosine similarity, a
pooled-vector rather than token-level comparison, does not have this
problem and separated relevant from irrelevant pairs in the same spot
checks ($\sim$0.65--0.75 vs.\ $-$0.10 to 0.25). A response counts as
grounded if its maximum ROUGE-1 recall across retrieved chunks reaches
0.15; cosine is reported as a continuous complement rather than a second
threshold.

Because generation depends on a non-deterministic local model, the
evaluation was repeated three times. The ROUGE-1 groundedness rate varied
across runs (10.0\%, 25.0\%, 35.0\%; mean 23.3\%) while retrieval, topic
coherence, and detection accuracy were stable.
Table~\ref{tab:groundedness} therefore reports the rate as a mean with
range; the continuous metrics and the per-query scores in
Table~\ref{tab:perquery} are taken from the highest-scoring run and
should be read as an upper bound on that run-to-run distribution.

\begin{table}[htbp]
\centering
\small
\caption{Groundedness across 20 held-out queries. The groundedness rate is
the mean over three generation runs, with the observed range in brackets;
continuous metrics are from a single run. City accuracy is query-side city
detection; feature accuracy is detection of at least one expected aspect
category.}
\label{tab:groundedness}
\begin{tabular}{@{}lc@{}}
\toprule
\textbf{Metric} & \textbf{Score} \\
\midrule
Groundedness rate (ROUGE-1 $\geq$ 0.15) & 23.3\% [10.0--35.0] \\
Mean / median ROUGE-1 recall            & 0.139 / 0.135 \\
Mean / median cosine similarity         & 0.730 / 0.727 \\
Query-side city detection accuracy      & 100.0\% (20/20) \\
Feature detection accuracy              & 80.0\% (16/20) \\
\bottomrule
\end{tabular}
\end{table}

\begin{table}[htbp]
\centering
\footnotesize
\setlength{\tabcolsep}{6pt}
\caption{Per-query ROUGE-1 recall and sentence-embedding cosine
similarity, from the highest-scoring of three generation runs.
$\checkmark$ marks queries meeting the ROUGE-1 threshold of 0.15.}
\label{tab:perquery}
\begin{tabular}{@{}lccc@{}}
\toprule
\textbf{Query} & \textbf{ROUGE-1} & \textbf{Cosine} & \textbf{Grounded} \\
\midrule
What do people say about road safety in Charlotte? & 0.206 & 0.777 & \checkmark \\
What is housing like in Montreal?                  & 0.172 & 0.716 & \checkmark \\
Are there good bike lanes in Seattle?              & 0.169 & 0.778 & \checkmark \\
How dense is housing in Montreal?                  & 0.166 & 0.722 & \checkmark \\
How do people feel about car dependency in Houston? & 0.155 & 0.721 & \checkmark \\
How is cycling infrastructure in Portland?         & 0.153 & 0.751 & \checkmark \\
What do people say about pedestrian safety in Manhattan? & 0.151 & 0.707 & \checkmark \\
\midrule
Is Charlotte walkable?                             & 0.150 & 0.710 & \\
Is New York City bikeable?                         & 0.146 & 0.776 & \\
Is Miami walkable?                                 & 0.139 & 0.770 & \\
How is cycling in Montreal?                        & 0.130 & 0.815 & \\
What do people think about transit in Miami?       & 0.124 & 0.654 & \\
How is walkability in Seattle?                     & 0.123 & 0.759 & \\
What do people say about parking in Miami?         & 0.122 & 0.669 & \\
How safe are the streets in Miami?                 & 0.122 & 0.639 & \\
Is Miami transit good or bad?                      & 0.120 & 0.766 & \\
Is Edmonton bikeable?                              & 0.117 & 0.766 & \\
What urban design issues exist in Houston?         & 0.117 & 0.733 & \\
Is Houston car dependent?                          & 0.106 & 0.689 & \\
What are transit options in Portland?              & 0.098 & 0.686 & \\
\bottomrule
\end{tabular}
\end{table}

Query-side city detection was correct on all 20 queries. This measures
parsing of short, well-formed questions, and is not evidence about
corpus-side resolution over noisy comments. Feature detection reached
80.0\%, failing on four queries: a housing query that missed
\texttt{density\_signal}, two safety queries where ``street safety'' and
``pedestrian safety'' did not match \texttt{safety}, and an urban-design
query that missed \texttt{walkability}.

\section{Discussion}

\subsection{Why Both Sentiment Models Fail on Neutral}
Both failure modes trace to domain mismatch with a register neither model
was built for---traffic- and pedestrian-safety commentary.

VADER over-predicts \textit{positive}: 37 of 103 negative comments
(35.9\%) and 61 of 122 neutral comments (50.0\%) are called positive,
driving neutral recall to 24.6\%. Dashcam and traffic-safety comment
culture uses superficially positive words (``nice'', ``great job'')
sarcastically, which VADER scores at face value, and domain-specific
negative vocabulary (\textit{stroad}, \textit{car-centric}) is absent from
its dictionary.

The transformer keeps strong negative recall (79.6\%) but splits neutral
errors roughly evenly between positive (24.6\% of neutrals) and negative
(29.5\%), yielding 45.9\% neutral recall. Urbanist comments carry
substantial analytical and informational content---geographic comparisons,
historical context, factual corrections about a crash or
intersection---that the model reads as polarized. The positive class is
weakest for both, for opposite reasons: VADER inflates positive recall to
76.0\% at the cost of precision, while the transformer loses a third of
true positives to neutral. Urbanist praise is also domain-specific
(``so easy to live without a car''), and neither model is calibrated for
it. The annotator disagreement pattern ($\kappa = 0.53$, concentrated on
the positive class) suggests part of this is intrinsic class ambiguity in
the register rather than model error alone.

For city profile construction the transformer's negative recall
(79.6\% vs.\ 52.4\%) is the consequential difference: criticism of car
dependency, poor transit, and dangerous streets is precisely the signal
the profiles depend on.

\subsection{ROUGE-1 Groundedness Is a Paraphrase Floor}
ROUGE-1 clusters in a narrow band (mean 0.139, median 0.135, range
0.098--0.206), with most sub-threshold queries just below 0.15 rather than
near zero---already suggestive of paraphrase rather than hallucination.
Every query scores cosine between 0.639 and 0.815 (mean 0.730), within the
range spot checks established for relevant pairs ($\sim$0.65--0.75) and
far above the range for irrelevant pairs ($-$0.10 to 0.25). Even the four
lowest-ROUGE queries---Portland transit (0.098/0.686), Houston car
dependency (0.106/0.689), Houston urban design (0.117/0.733), Edmonton
bikeability (0.117/0.766)---sit inside that range
(Table~\ref{tab:perquery}). The two metrics
correlate only moderately ($r = 0.34$), as expected: ROUGE-1 penalises the
paraphrase and reordering cosine is invariant to. The ROUGE-1 groundedness
rate is therefore better read as a lower bound on faithfulness than as a
measure of how often responses depart from their evidence.

Generation quality also bounds the result. Llama 3.2 was chosen for cost
and local availability, not maximum quality; a larger instruction-tuned
model would plausibly track evidence more closely at the surface level,
raising ROUGE-1 with no change to retrieval or data.

\subsection{Keyword Taxonomy Coverage}
The four feature-detection misses show the fixed taxonomy does not cover
natural phrasings: ``housing'' missed \texttt{density\_signal}, ``street
safety'' and ``pedestrian safety'' missed (or, for Manhattan, matched
\texttt{walkability} instead of) \texttt{safety}, and ``urban design
issues'' missed \texttt{walkability}. These reduce the precision of the
aspect re-ranking step and can degrade the relevance of retrieved
documents, though the cosine scores across all 20 queries suggest this did
not meaningfully harm groundedness for this query set.

\section{Conclusion and Future Work}

We presented a pipeline that turns noisy urbanist YouTube discourse into
queryable, geographically grounded, per-city sentiment profiles and a RAG
conversational interface over them, and reported where standard NLP
components break in this setting.

Several directions follow directly. Fine-tuning the sentiment classifier
on 2{,}000--5{,}000 domain-labeled examples---applying the
topic-versus-video rule and prioritised by active learning---should
address the 45.9\% neutral recall that most limits profile quality. The
evaluation would be strengthened by an ablation isolating the contribution
of city filtering and aspect re-ranking, an LDA reference point for topic
coherence, a manual audit of corpus-side city assignment, a random-pairing
control for the cosine metric, and chunk-level relevance rating of both
retrieval systems under a blind, multiply-rated protocol. Deterministic
decoding, or averaging over more generation runs, would remove the
run-to-run variance in the groundedness rate, and a stronger
instruction-tuned model would separate generation quality from retrieval
quality. The keyword taxonomy could be replaced with zero-shot NLI
classification to generalise to unseen phrasings, and geographic
resolution would benefit from a geoparser with a containment hierarchy and
global coverage. Finally, incorporating dated structured sources---GTFS
feeds, Mapillary imagery, Walk Score---would ground profiles in timestamped
data and address the pipeline's inability to detect comments describing
outdated conditions.

\section{Limitations}
\label{sec:limitations}

The 300-comment sentiment validation set is small relative to the corpus
and was labeled by two annotators rather than a panel, with moderate
agreement ($\kappa = 0.53$) concentrated on the positive class.

Groundedness is measured by ROUGE-1 recall and embedding cosine, which
correlate only moderately ($r = 0.34$) and are proxies for, not measures
of, factual correctness. Because generation depends on a non-deterministic
local model, the groundedness rate varied across repeated runs
(10.0\%, 25.0\%, 35.0\%) even though retrieval, topic coherence, and
detection accuracy remained stable; the reported figures are indicative
rather than exact. Some queries fell back to the template-based generator
when the model server was unavailable. Because that generator copies
evidence spans directly, it inflates ROUGE-1 recall relative to
LLM-generated text, and the two are not separated in the reported figures.

Retrieval relevance judgments are auto-suggested at the video level and
validated by a single-rater, non-blind chunk-level rating of the dense
system only; no comparable rating exists for the TF-IDF baseline. We
report no ablation isolating city filtering and aspect re-ranking, so the
contribution of these components over plain dense retrieval is not
quantified, and the 0.2 aspect weight and the fuzzy-match threshold of 95
are not tuned. Topic coherence is reported without a within-corpus
baseline. Cosine similarity is interpreted against ranges established by
manual spot checks rather than a random-pairing control, so it may partly
reflect domain similarity rather than groundedness.

Corpus-side geographic resolution is not evaluated: query-side city
detection is perfect on 20 short, well-formed questions, but the city
assignments underlying the profiles are not audited. Geographic coverage
is limited to a US/Canada gazetteer. The evaluation set originally
included Stockholm and London; both resolved to same-named North American
towns and were replaced. The same mechanism means the ``London'' row of
Table~\ref{tab:cityprofiles} conflates mentions of London, UK with a North
American entry, and New York and Manhattan are not merged. The
topic-relevance classifier discards 45\% of the length-filtered corpus and
is not independently evaluated, so its error rate is unknown and
propagates to every downstream result.

The corpus is drawn from urbanist-focused channels, biasing toward an
audience already engaged with urban planning rather than the general
public, and is English-only, so the findings do not speak to multilingual
or code-switched urbanist discourse. It has no temporal grounding, so
comments may describe conditions that no longer hold.

\section{Ethics Statement}

\paragraph{Data collection} All data were collected through two official
interfaces: the YouTube Data API v3 (\texttt{commentThreads} endpoint) for
user comments, and the \texttt{youtube-\allowbreak transcript-\allowbreak api} Python library for
video transcripts, which reads only transcripts a creator has made publicly
available. Both access content visible to any unauthenticated viewer. No
private data, no authenticated endpoints, no login credentials, and no
scraping of YouTube's HTML interface were used at any stage, and no fallback
scraper or proxy service was employed for videos the transcript library
failed on. All requests were made through documented, rate-limited API calls
under a standard Google Cloud project, consistent with the YouTube API
Services Terms of Service. Videos were selected purely by keyword search;
we did not filter by channel or license, and we collected no private
messages, age-restricted content, or material from private or unlisted
videos.

\paragraph{Human subjects review} This work analyses publicly posted
content and models no author-level attribute, and was determined not to
constitute human subjects research under 45 CFR 46. We nonetheless applied
the data-minimisation practices described below rather than treating public
availability as sufficient justification on its own.

\paragraph{Personal information} YouTube comments are authored by
identifiable accounts. Only comment text, video identifier, and content type
(comment or transcript) were retained at collection time; usernames, channel
identifiers, profile information, and all other author-level fields were
discarded rather than stored and later stripped. No author-level attribute is
modeled, inferred, or reported anywhere in this work, and all analysis is
aggregated to the city level. No attempt was made to identify, profile, or
link any individual commenter.

\paragraph{Release} Consistent with the platform's terms of service, any
release of this corpus would consist of video and comment identifiers
together with derived annotations---resolved city, aspect labels, and
sentiment label---rather than comment text, allowing rehydration by other
researchers.

\paragraph{Intended use and misuse} City sentiment profiles derived from
a self-selected urbanist audience are not representative of city residents
and should not be read as public opinion measurement. We report this
selection bias explicitly in Section~\ref{sec:limitations} to discourage
such use. The profiles are descriptive of online discourse only, and
should not inform policy decisions without independent, representative
data.

\paragraph {\textbf{AI Use Statement}}
ChatGPT was used to improve language and clarity and to assist with creating and formatting tables. Claude was used for formatting assistance. The authors reviewed, revised, and approved all AI-assisted content. No AI tool was used to generate, alter, or interpret the research data, results, analysis, or conclusions.

\bibliographystyle{elsarticle-harv}
\bibliography{custom}

@misc{walkscore_methodology,
  author       = {{Walk Score}},
  title        = {Walk Score Methodology},
  howpublished = {\url{https://www.walkscore.com/methodology.shtml}},
  note         = {Accessed: Feb. 2026}
}

@article{manaugh2011validating,
  author  = {Manaugh, Kevin and El-Geneidy, Ahmed},
  title   = {Validating Walkability Indices: How Do Different Households Respond to the Walkability of Their Neighbourhood?},
  journal = {Transportation Research Part D: Transport and Environment},
  volume  = {16},
  number  = {4},
  pages   = {309--315},
  year    = {2011}
}

@inproceedings{nahas2024classifiers,
  author    = {Nahas, Naeema and Swetha, P. and Nandakumar, R.},
  title     = {Classifiers for Sentiment Analysis of {YouTube} Comments: A Comparative Study},
  booktitle = {Proceedings of the International Conference on Machine Learning, Deep Learning and Computational Intelligence for Wireless Communication (MDCWC 2023)},
  series    = {Signals and Communication Technology},
  publisher = {Springer},
  address   = {Cham},
  pages     = {443--451},
  year      = {2024},
  doi       = {10.1007/978-3-031-47942-7_38}
}

@inproceedings{hutto2014vader,
  author    = {Hutto, Clayton J. and Gilbert, Eric},
  title     = {{VADER}: A Parsimonious Rule-Based Model for Sentiment Analysis of Social Media Text},
  booktitle = {Proceedings of the 8th International AAAI Conference on Weblogs and Social Media (ICWSM-14)},
  pages     = {216--225},
  year      = {2014},
  address   = {Ann Arbor, MI}
}

@article{grootendorst2022bertopic,
  author  = {Grootendorst, Maarten},
  title   = {{BERTopic}: Neural Topic Modeling with a Class-Based {TF-IDF} Procedure},
  journal = {arXiv preprint arXiv:2203.05794},
  year    = {2022}
}

@article{zapletal2023social,
  author  = {Zapletal, Filip and others},
  title   = {Social Media, Topic Modeling and Sentiment Analysis in Municipal Decision Making},
  journal = {arXiv preprint arXiv:2308.04124},
  year    = {2023}
}

@inproceedings{lewis2020rag,
  author    = {Lewis, Patrick and Perez, Ethan and Piktus, Aleksandra and Petroni, Fabio and Karpukhin, Vladimir and Goyal, Naman and K{\"u}ttler, Heinrich and Lewis, Mike and Yih, Wen-tau and Rockt{\"a}schel, Tim and Riedel, Sebastian and Kiela, Douwe},
  title     = {Retrieval-Augmented Generation for Knowledge-Intensive {NLP} Tasks},
  booktitle = {Advances in Neural Information Processing Systems},
  volume    = {33},
  pages     = {9459--9474},
  year      = {2020}
}

@article{gao2023ragsuvey,
  author  = {Gao, Yunfan and others},
  title   = {Retrieval-Augmented Generation for Large Language Models: A Survey},
  journal = {arXiv preprint arXiv:2312.10997},
  year    = {2023}
}

@article{zhang2020bertscore,
  author  = {Zhang, Tianyi and Kishore, Varsha and Wu, Felix and
             Weinberger, Kilian Q. and Artzi, Yoav},
  title   = {{BERTScore}: Evaluating Text Generation with {BERT}},
  journal = {arXiv preprint arXiv:1904.09675},
  year    = {2020}
}

@article{landis1977kappa,
  author    = {Landis, J. Richard and Koch, Gary G.},
  title     = {The Measurement of Observer Agreement for Categorical Data},
  journal   = {Biometrics},
  volume    = {33},
  number    = {1},
  pages     = {159--174},
  year      = {1977},
  publisher = {International Biometric Society},
  doi       = {10.2307/2529310}
}

@inproceedings{sanh2019distilbert,
    title     = {{DistilBERT}, a distilled version of {BERT}: smaller, faster, cheaper and lighter},
    author    = {Sanh, Victor and Debut, Lysandre and Chaumond, Julien and Wolf, Thomas},
    booktitle = {5th Workshop on Energy Efficient Machine Learning and Cognitive Computing (EMC$^2$), NeurIPS},
    year      = {2019}
}

@inproceedings{reimers2019sbert,
    title     = {{Sentence-BERT}: Sentence Embeddings using {S}iamese {BERT}-Networks},
    author    = {Reimers, Nils and Gurevych, Iryna},
    booktitle = {Proceedings of the 2019 Conference on Empirical Methods in Natural Language Processing and the 9th International Joint Conference on Natural Language Processing (EMNLP-IJCNLP)},
    pages     = {3982--3992},
    year      = {2019}
}

@article{johnson2019faiss,
    title   = {Billion-scale similarity search with {GPU}s},
    author  = {Johnson, Jeff and Douze, Matthijs and J{\'e}gou, Herv{\'e}},
    journal = {IEEE Transactions on Big Data},
    volume  = {7},
    number  = {3},
    pages   = {535--547},
    year    = {2019}
}

@misc{honnibal2020spacy,
    title        = {{spaCy}: Industrial-strength Natural Language Processing in Python},
    author       = {Honnibal, Matthew and Montani, Ines and Van Landeghem, Sofie and Boyd, Adriane},
    year         = {2020},
    doi          = {10.5281/zenodo.1212303},
    howpublished = {Zenodo}
}

@article{mcinnes2017hdbscan,
    title   = {hdbscan: Hierarchical density based clustering},
    author  = {McInnes, Leland and Healy, John and Astels, Steve},
    journal = {Journal of Open Source Software},
    volume  = {2},
    number  = {11},
    pages   = {205},
    year    = {2017}
}

@inproceedings{lin2004rouge,
    title     = {{ROUGE}: A Package for Automatic Evaluation of Summaries},
    author    = {Lin, Chin-Yew},
    booktitle = {Text Summarization Branches Out},
    pages     = {74--81},
    year      = {2004},
    publisher = {Association for Computational Linguistics}
}

@article{liu2019roberta,
    title   = {{RoBERTa}: A Robustly Optimized {BERT} Pretraining Approach},
    author  = {Liu, Yinhan and Ott, Myle and Goyal, Naman and Du, Jingfei and Joshi, Mandar and Chen, Danqi and Levy, Omer and Lewis, Mike and Zettlemoyer, Luke and Stoyanov, Veselin},
    journal = {arXiv preprint arXiv:1907.11692},
    year    = {2019}
}

@inproceedings{loureiro2022timelms,
    title     = {{TimeLMs}: Diachronic Language Models from {T}witter},
    author    = {Loureiro, Daniel and Barbieri, Francesco and Neves, Leonardo and Espinosa Anke, Luis and Camacho-Collados, Jose},
    booktitle = {Proceedings of the 60th Annual Meeting of the Association for Computational Linguistics: System Demonstrations},
    pages     = {251--260},
    year      = {2022}
}

@article{grattafiori2024llama3,
    title   = {The {Llama} 3 Herd of Models},
    author  = {Grattafiori, Aaron and Dubey, Abhimanyu and Jauhri, Abhinav and others},
    journal = {arXiv preprint arXiv:2407.21783},
    year    = {2024}
}

@inproceedings{gritta2018melbourne,
    title     = {Which {M}elbourne? Augmenting Geocoding with Maps},
    author    = {Gritta, Milan and Pilehvar, Mohammad Taher and Collier, Nigel},
    booktitle = {Proceedings of the 56th Annual Meeting of the Association for Computational Linguistics (Volume 1: Long Papers)},
    pages     = {1285--1296},
    year      = {2018}
}

@article{gritta2020pragmatic,
    title   = {A Pragmatic Guide to Geoparsing Evaluation},
    author  = {Gritta, Milan and Pilehvar, Mohammad Taher and Collier, Nigel},
    journal = {Language Resources and Evaluation},
    volume  = {54},
    number  = {3},
    pages   = {683--712},
    year    = {2020}
}

@article{halterman2017mordecai,
    title   = {Mordecai: Full Text Geoparsing and Event Geocoding},
    author  = {Halterman, Andrew},
    journal = {Journal of Open Source Software},
    volume  = {2},
    number  = {9},
    pages   = {91},
    year    = {2017}
}

@book{leidner2008toponym,
    title     = {Toponym Resolution in Text: Annotation, Evaluation and Applications of Spatial Grounding of Place Names},
    author    = {Leidner, Jochen L.},
    year      = {2008},
    publisher = {Universal Press},
    address   = {Boca Raton, FL}
}

@misc{gtrends_urbanism,
  author       = {{Google Trends}},
  title        = {Search interest in {``walkability''}, {``public transit''}, {``mixed-use''}, {``housing density''}, and {``safe streets''}, United States, 2004--2026},
  year         = {2026},
  howpublished = {\url{https://trends.google.com/trends/explore}},
  note         = {Accessed: Feb. 2026}
}

\appendix

\section{Implementation Details}
\label{app:impl}

The system is implemented in Python 3.13 using spaCy for NER, rapidfuzz
for fuzzy matching, BERTopic and sentence-transformers for topic modeling
and encoding, faiss-cpu for indexing, transformers and torch for sentiment
classification, Flask 3.1 for the REST API and web interface, and
pandas/numpy for data processing. The dataset is loaded from CSV and
processed in memory at startup, including deduplication and derived fields
for resolved city, aspects, and sentiment. The FAISS index and BERTopic
model are built over the full corpus at initialization and held in memory.
The backend exposes REST endpoints for query handling, raw retrieval
inspection, and city profile retrieval. The active language model backend
is auto-detected from the runtime environment. Everything runs on CPU;
BERTopic fitting and index construction over 22{,}788 chunks take several
minutes on a standard laptop.

\end{document}